\documentclass{article}

\usepackage{microtype}
\usepackage{graphicx}
\usepackage{subcaption}
\usepackage{booktabs} % for professional tables

\usepackage{hyperref}

\usepackage[accepted]{icml2026}

\usepackage{amsmath}
\usepackage{amssymb}
\usepackage{mathtools}
\usepackage{amsthm}

\usepackage{multirow}
\usepackage{xcolor}
\definecolor{darkgreen}{RGB}{0,128,0}
\newcommand{\COMMENTSTATE}[1]{\color{darkgreen} \STATE \# {#1} \color{black}}

\usepackage[capitalize,noabbrev]{cleveref}

\theoremstyle{plain}

\theoremstyle{definition}

\theoremstyle{remark}

\usepackage[textsize=tiny]{todonotes}
\icmltitlerunning{EasyBalance: Cross-Layer Load Balancing in Distributed MoE Inference}

\begin{document}

\twocolumn[
  \icmltitle{EasyBalance: Cross-Layer Load Balancing in Distributed MoE Inference}



  \icmlsetsymbol{equal}{*}

  \begin{icmlauthorlist}
    \icmlauthor{Yize Wu}{isrc,ucas}
    \icmlauthor{Ke Gao}{isrc}
    \icmlauthor{Ling Li}{isrc,ucas}
    \icmlauthor{Yanjun Wu}{isrc}
  \end{icmlauthorlist}

  \icmlaffiliation{isrc}{Intelligent Software Research Center, Institute of Software, CAS, Beijing, China}
  \icmlaffiliation{ucas}{University of Chinese Academy of Sciences, Beijing, China}

  \icmlcorrespondingauthor{Yanjun Wu}{yanjun@iscas.ac.cn}

  \icmlkeywords{Inference Acceleration, Mixture of Experts, Distributed Inference, Expert-Parallel Load Balancing}

  \vskip 0.3in
]



\printAffiliationsAndNotice{}  % no special notice (required even if empty)

\begin{abstract}
  Load Balancing has emerged as a critical problem in expert-parallel distributed inference of Mixture-of-Experts (MoE) models. As routing distributions are typically skewed across experts, devices hosting lighter-loaded experts must idle to wait for the heaviest during expert computing, leading to inefficiency. Existing load-balancing approaches primarily rely on expert replication or migration within each layer, which introduce additional overhead and limit their flexibility and scalability. To address this problem, we propose EasyBalance, a \textbf{cross-layer} load balancing strategy that requires no modifications to the expert-device mapping, enabling instant adaptability and incurring essentially no additional overhead. Our key insights are that (1) experts of other layers can be viewed as naturally redundant for the current layer, and (2) cross-layer MoE workloads can be jointly executed to mitigate their individual imbalance. Based on these observations, EasyBalance greedily schedules a subset of cross-layer workloads to run at each MoE step and defers the remaining workloads for future balancing opportunities, effectively leveraging cross-layer imbalance mitigation. Extensive experiments across models, tasks, and configurations demonstrate that EasyBalance consistently accelerates distributed MoE inference, reducing GPU idling by mostly over 40\%. Code is available at https://github.com/yize-wu/EasyInfra.
\end{abstract}

\begin{figure*}[t]
    \centering
    \includegraphics[width=0.9\linewidth]{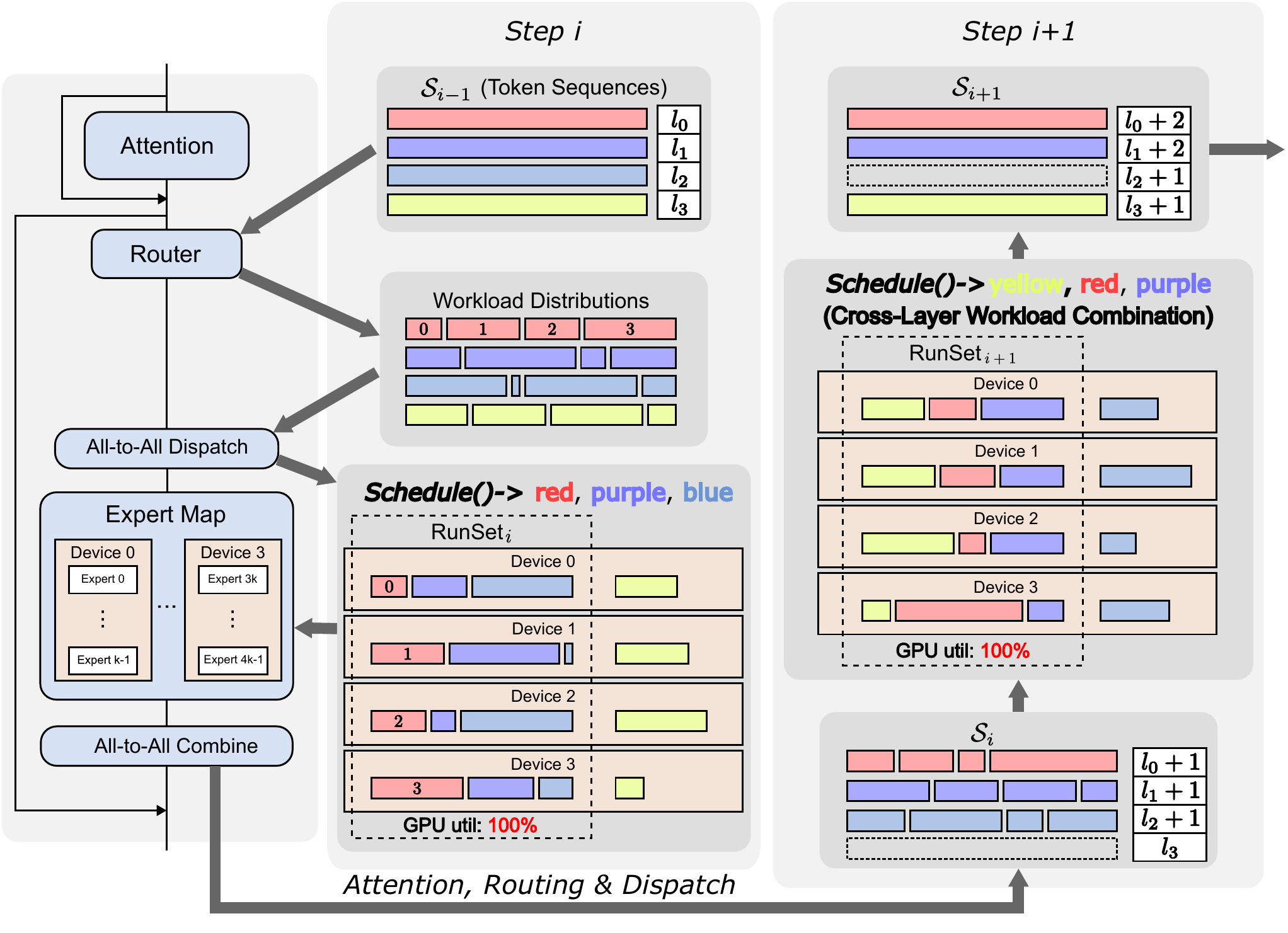}
    \caption{Demonstration of EasyBalance. MoE workloads of micro-batches (indicated by colors) can be selectively scheduled and jointly executed while residing at different layers, achieving improved GPU utilization. $l_j$ indicates the current layer of micro-batch $j$.}
    \label{fig:overview}
\end{figure*}

\section{Introduction}

The rapidly growing parameter sizes of transformer-based models \cite{transformer} present substantial computational challenges \cite{training_large}. Mixture-of-Expert (MoE) architecture \cite{MoE} offers a solution by replacing dense feed-forward layers with multiple smaller expert networks, where each token activates only a sparse subset of them. This sparse activation mechanism significantly reduces per-token computation cost \cite{switch_transformers}, while enabling models with substantially larger overall parameter counts.

In distributed MoE inference, experts are typically dispatched across multiple devices to parallelize their computation workloads, known as expert parallelism. In this setting, when tokens are routed to experts residing on remote devices, the tensors of their hidden representations are first dispatched to the corresponding devices via an all-to-all communication. Then, after expert computation, a second all-to-all gathers the results for aggregation \cite{toward_efficient, tutel}. 

While expert parallelism can substantially accelerate MoE inference, load imbalance has emerged as a critical performance bottleneck \cite{deepseekv3}. Despite the use of load-balancing auxiliary loss during MoE training, inference-time token routing often remains uneven across experts, leading to skewed computational workloads among devices. All the devices must wait for the most heavily loaded one to finish its computation, as the operations of result gathering and aggregation require synchronization. This results in severe resource under-utilization and system inefficiency \cite{netmoe}.

Existing methods for mitigating load balancing primarily rely on expert replication or migration within each expert layer\cite{lina, eplb, harmoeny}. By modifying the expert–device mapping of MoE models in the system, these methods attempt to redistribute workloads more evenly across devices. Despite their effectiveness, such approaches suffer from two inherent limitations. First, they lack flexibility when serving newly assigned tasks, as expert routing patterns can vary substantially across inputs and applications (see \cref{fig:routing_distribution}). Second, expert replication incurs considerable memory overhead, while expert migration introduces additional communication costs, both of which pose limitation on scalability, which is a critical concern as load imbalance becomes increasingly severe with larger-scale deployments (as demonstrated in \cref{tab:ablation_ep_size}).

To overcome these limitations, we propose EasyBalance, a novel cross-layer load balancing strategy. Unlike prior approaches, EasyBalance requires no changes to the expert–device mapping throughout the inference process, thereby providing instant task adaptability with essentially no additional memory or communication overhead. EasyBalance is motivated by two key observations:

\textbf{Cross-Layer Expert Redundancy.} Expert replication \cite{eplb} creates load-balancing opportunities by introducing redundant experts, but at the cost of increased memory consumption. However, from a novel cross-layer perspective, we observe that redundancy already exists in an inference system: for a given expert layer, experts of other layers can be regarded as naturally ``redundant'‘, as they already reside in device memory (just like the replication). These experts can therefore be leveraged to balance workloads without introducing any additional overhead, yet this potential remains largely unexplored.

\textbf{Cross-Layer Workload Combination.} While model inference must follow a strict layer-wise sequential order, different micro-batches can progress through multiple layers simultaneously, which allows MoE workloads from multiple layers to be jointly executed in one step. As shown in \cref{fig:overview}, the sequences can run on different MoE layers ($l_j$ and $l_k$ can be different for $j\ne k$, $j,k \in \{0,1,2,3\}$) with their workloads combined. Crucially, workload combination is worst-case performance-safe and mostly beneficial: (1) for two workload distributions $\boldsymbol{w}_1,\boldsymbol{w}_2 \in \mathbb{Z}^D$, the combined workload will be $\boldsymbol{w}_1+\boldsymbol{w}_2$. Since MoE computation is bottlenecked by the heaviest rank, (i.e., $\max(\boldsymbol{w})$), the actual system workloads would satisfy the worst-case safe condition of $\max(\boldsymbol{w}_1 + \boldsymbol{w}_2) \le \max(\boldsymbol{w}_1) + \max(\boldsymbol{w}_2)$. This property extends to multiple workloads as well; (2) the probability of encountering the worst case is low, as it only occurs when multiple workloads skew on the same device, and this probability even decreases as the number of devices scales up. Therefore, cross-layer workload combination is typically beneficial for modern deep models, particularly in large-scale distributed settings (see detailed explanation in \cref{subsec:method_cross_layer_perspective}, and empirical evidences in \cref{tab:compensate}).

Based on these observations, EasyBalance performs cross-layer scheduling over micro-batches to mitigate load imbalance within the whole inference process. At each MoE computation step, it selectively schedules a subset of workloads for execution and defers others to wait for potential future balancing opportunities, to leverage the aforementioned property of cross-layer imbalance mitigation. Various heuristic scheduling algorithms have all been proven effective in practice, demonstrating the robustness of our method. Experimental results across models and tasks demonstrate that EasyBalance can consistently accelerate the inference process by mitigating load imbalance, with GPU under-utilization reduced by mostly over 40\%.

\section{Preliminary}

\subsection{Mixture-of-Expert Architecture}

A transformer layer typically consists of two main modules: self-attention and multi-layer perceptron (MLP). Given the input hidden state $h_l$ of layer $l$, the output of each module is aggregated with residual connections \cite{resnet} as:
\[
    h'_{l} = h_l + \text{Attention}(h_l)
\]
\[
    h_{l+1} = h'_l + \text{MLP}(h'_l)
\]

The Mixture-of-Expert (MoE) architecture replaces the dense MLP with a set of experts, which are sparsely activated through a routing mechanism. For each token, the router selects the top-$k$ experts according to the gating function $G$:
\[
     W_l, I_l = \text{topk}(G(h'_l)).
\]

Each selected expert processes the token independently, and their outputs are aggregated using a weighted sum:

\begin{equation}
    MLP(h'_l) = \sum_{i \in I_l} W_{l,i} * \text{expert}_i(h'_l),
\label{eq:weighted_aggregation_of_moe}
\end{equation}

where $W_{l,i}$ denotes the routing weight for expert $i$. The aggregated result is then passed to the next layer as the output of the MoE module.

\subsection{Expert Parallelism}

In distributed MoE inference, expert parallelism (EP) is commonly adopted to dispatch expert computation across multiple devices. Each expert is placed onto one or more devices according to an expert–device mapping, and each device hosts a subset of experts and processes tokens routed to them.

Under EP, tokens may be assigned to experts that reside on remote devices. To solve this, an all-to-all communication is required to dispatch token representations to the corresponding devices, and after MoE computation the expert outputs will be communicated back to the original devices via another all-to-all, where the weighted aggregation (\cref{eq:weighted_aggregation_of_moe}) is performed.

\begin{table}[t]
    \centering
    \begin{tabular}{c|cc}
        \toprule
        \multirow{2}{*}{Model} & \multicolumn{2}{c}{End-to-end Latency} \\
        \cmidrule(lr){2-3}
        & w/o micro-batch & w/ micro-batch (4 splits) \\
        \midrule
        Q3-30B & 10.15 & 8.60 \\
        M-16B & 5.06 & 4.40 \\
        \bottomrule
    \end{tabular}
    \caption{End-to-end latency(s) of Qwen3-30B (Q3-30B) and Moonlight-16B (M-16B) on 2wikimqa and EP=8. The batch size and sequence length are 128 and 4K.}
    \label{tab:micro_batch_ttft}
\end{table}

\subsection{Load Balancing}
\label{subsec:preliminary_load_balancing}

Although MoE training typically incorporates an auxiliary load-balancing loss to encourage balanced token routing, the routing distribution at inference time still often remains skewed, leading to uneven workload distribution across devices under expert parallelism. As the gathering communication and weighted aggregation are both synchronous operations, devices with lighter workloads must idle while waiting for the most heavily-loaded one to finish, introducing hardware under-utilization and therefore suboptimal performances.

Formally, denote the MoE computation workload at layer $l$ across $D$ devices as $\boldsymbol{w}^{(l)}=(w_1^{(l)},\dots,w_D^{(l)})$. Distributed MoE computation is bottlenecked by the slowest device, so the effective workload will be:
\begin{equation}
    \hat{w}^{(l)} = \max(\boldsymbol{w}^{(l)}),
\label{eq:effective_workload_def}
\end{equation}
which is directly proportional to execution latency. Moreover, the overall device utilization across all the layers is defined as:
\begin{equation}
    u = \frac{\sum_{l=1}^{l_{max}}\sum_{i=1}^D w^{(l)}_i}{\sum_{l=1}^{l_{max}}\hat{w}^{(l)} \times D},
\label{eq:utilization_def}
\end{equation}
which quantifies hardware efficiency by measuring the ratio of used resources to total available capacity.

\subsection{Micro-Batching}

Micro-batching is a widely used technique in large-scale inference scenarios. It splits a large batch of input sequences into multiple micro-batches and pipelines their execution \cite{vllm, sglang}. The benefits include computation-communication overlapping \cite{step3}, efficient memory footprint, KV cache transfer in prefilling-decoding disaggregation \cite{zcube}, etc.. For example, empirical results in \cref{tab:micro_batch_ttft} demonstrate that micro-batching can significantly reduce end-to-end inference latency by hiding the communication overhead behind other batches' computation. EasyBalance builds upon micro-batching to further enable cross-layer workload balancing during MoE computation.

\section{Problem}
\label{sec:problem}

Existing approaches address load imbalance through expert replication and/or migration within each MoE layer, to achieve more balanced workload redistribution across devices. Specifically, experts replication allow tokens routed to ``hot'' experts to be selectively dispatched across multiple copies, while expert migration alleviates device hotspots by relocating hot experts to other less-loaded GPUs. These techniques typically require modifying the expert-device mapping based on routing-distribution statistics.

Despite their effectiveness, current solutions suffer from two inherent limitations that restrict their applicability in broader inference scenarios:

\textbf{Flexibility}. MoE routing patterns are highly task-dependent (as shown in \cref{fig:routing_distribution}), and an effective expert mapping for a specific task may become suboptimal---or even harmful---for inputs from other tasks. Therefore, the reliance on task- or data-specific information fundamentally limits the flexibility.

\textbf{Scalability}. Updating the expert mapping incurs non-negligible overheads. Expert replication introduces substantial memory consumption, while expert migration incurs additional communication and synchronization costs. These overheads scale positively with both model sizes and distribution scales, limiting the application in large-scale settings.

The limitations motivate the need for a fundamentally different load-balancing method—one that does not require modification of the expert mapping and remains both flexible and scalable across diverse inference scenarios.

\begin{table}[t]
    \centering
    \begin{tabular}{c|cc}
        \toprule
        \multirow{2}{*}{Task} & \multicolumn{2}{c}{$N$} \\
        \cmidrule(lr){2-3}
        ~ & 2 & 3 \\
        \midrule
        2wikimqa & 2/47 & 0/46 \\
        trec & 4/47 & 0/46 \\
        repobench-p & 10/47 & 1/46 \\
        all & 77/611=0.126 & 4/598=0.006 \\
        \bottomrule
    \end{tabular}
    \caption{Numbers of $N$-consecutive layers that have identical workload distribution skew. The tested model is Qwen3-30B (48 MoE layers), with 32 batches and 4K sequence length. ``all'' stands for the statistics from all 13 representative tasks in LongBench.}
    \label{tab:compensate}
    % \vspace{-20pt}
\end{table}

\begin{algorithm}[t]
\caption{EasyBalance}
\begin{algorithmic}
\STATE \textbf{Input:} Micro-batches of tokens $T_0,\dots,T_{N-1}$, minimum combination size $m$.
\COMMENTSTATE{Initialize the variables.}
\STATE $l_j ~\leftarrow 0$ \text{for} $j\in\{0,\dots,N-1\}$
\STATE $\mathcal{S}_0$ $\leftarrow \{T_0, \dots, T_{N-1}\}$
\STATE RunSet$_0$ $\leftarrow \{T_0, \dots, T_{N-1}\}$
\STATE $i \leftarrow 0$ 
\COMMENTSTATE{The loop continues until all batches finish.}
\WHILE{$|\mathcal{S}_i| \ne 0$}
    \COMMENTSTATE{Increase the step.}
    \STATE $i \leftarrow i+1$
    \COMMENTSTATE{Run other operations of last-step executed batches.}
    \STATE \textit{Attn\_Route\_Dispatch}(RunSet$_{i-1}$)
    \COMMENTSTATE{Choose a $m$-minimum combination.}
    \STATE RunSet$_{i}$ $\leftarrow$ \textit{Schedule}($\mathcal{S}_{i-1}$, $m$)
    \COMMENTSTATE{Execute the scheduled workload.}
    \STATE \textit{MoECompute\_Combine}(RunSet$_{i}$)
    \COMMENTSTATE{Increase $l_j$ of scheduled batches.}
    \FOR{$T_j \in $ RunSet$_{i}$}
        \STATE $l_j$ $\leftarrow$ $l_j + 1$
    \ENDFOR
    \COMMENTSTATE{Exclude finished batches.}
    \STATE $\mathcal{S}_{i} \leftarrow \{T_j| T_j \in \mathcal{S}_{i-1}, l_j < l_{max}\}$
\ENDWHILE

\end{algorithmic}
\label{alg:easybalance}
\end{algorithm}

\section{Method}

To address the aforementioned limitations, we propose EasyBalance, a cross-layer load balancing strategy for distributed MoE inference acceleration. EasyBalance requires no modification to the expert mapping, thereby offering instant flexibility and superior scalability with essentially no additional overheads.

\subsection{The Cross-Layer Perspective}
\label{subsec:method_cross_layer_perspective}

As discussed in \cref{sec:problem}, expert redundancy within a single MoE layer can only be achieved by expert replication, which inevitably incurs additional memory consumption. However, from a cross-layer perspective, we observe that redundancy inherently exists in the inference system: experts of other layers can be naturally considered as “redundant” for the current layer. Since these experts are already resident in GPU memory and ready for immediate usage without introducing any overhead, this observation suggests a promising solution to the current limitations.

Such potential has been overlooked previously, mainly due to the constraint of layer-wise sequential execution order of models, which prevents experts of other layers from directly participating in the computation. However, we observe that there are no dependencies among multiple micro-batches, making it feasible for micro-batches to simultaneously reside at different MoE layers. As illustrated in \cref{fig:overview}, the current layers of 4 micro-batches ($l_j$ for $j\in \{0,1,2,3\}$)) can differ from each other, while the inference process remains lossless as long as each micro-batch follows the sequential order of itself (each $l_j$ increments by 1 after one MoE computation of micro-batch $j$). The workloads of batches can be jointly executed (by multiple launches of their kernels) in one step, regardless of whether they are on the same layer.

Interestingly, cross-layer workload combination is worst-case performance-safe and highly likely to create load-balancing opportunities. 

\textbf{Worst-case safe.} According to \cref{eq:effective_workload_def}, the effective workload of combining multiple workloads $\boldsymbol{w}_j$ is $\max(\sum_j \boldsymbol{w}_{j})$, whereas executing them separately would result in $\sum_j\max(\boldsymbol{w}_{j})$. Since we always have $\max(\sum_j \boldsymbol{w}_{j})\le\sum_j\max(\boldsymbol{w}_{j})$, the combination never leads to degradation. 

\textbf{Mostly effective}. The worst case (when equality is achieved) only occurs when the workloads peak at the same device (i.e., $\text{argmax}_d(\boldsymbol{w}_j)$ is identical for all $j$), while its probability decreases when the number of devices grows (workloads are increasingly likely to peak on different devices with a large $D$). In other cases, the combination can reduce the effective workload due to strict inequality that $\max(\sum_j \boldsymbol{w}_{j})<\sum_j\max(\boldsymbol{w}_{j})$, accelerating MoE computation. More specifically, the imbalance of each workload $\boldsymbol{w}_j$ is ``compensated'' by those $\boldsymbol{w}_k$ with $\text{argmax}_d(\boldsymbol{w}_k) \ne \text{argmax}_d(\boldsymbol{w}_j)$. 

The empirical statistics of worst-case occurrences are reported in \cref{tab:compensate}, and the effectiveness of cross-layer workload combination is further demonstrated in \cref{fig:eplb_step_plot} and \cref{fig:more_eplb_step_plot}.

\subsection{Scheduling}
\label{subsec:method_batch_scheduling}

Based on cross-layer expert redundancy and workload combination, EasyBalance mitigates load imbalance through micro-batch scheduling. The end-to-end procedure is demonstrated in \cref{alg:easybalance}. 

At each MoE-computation step, the scheduler selects a subset of micro-batches (Runset$_i$) from the pool ($\mathcal{S}_{i-1}$) and executes the combined workload, while unselected batches are deferred and remain in $\mathcal{S}_{i}$ for future scheduling. Among the scheduled batches, those that have not reached the final layer (i,e., $l_j<l_{max}$) will advance to the next layer, progressing through attention, routing and all-to-all dispatching, and become ready for the next scheduling step. This scheduling mechanism allows micro-batches to reside on different layers during the procedure, and cross-layer imbalance mitigation is leveraged. The iterative process continues until all batches complete their execution at the last layer.

As the workload distributions of future layers are unknown at the current layer, the scheduling algorithm must be greedy. We explore several heuristic strategies, and discover that selecting the subset that maximizes GPU utilization genenrally yields the best performance (see \cref{subsec:scheduling_alg}). Additionally, to avoid executing overly small subsets---where computation and communication overlap becomes ineffective---we introduce a minimum threshold $m$ on the size of RunSet: the scheduling algorithm is required to return a new Runset with $|\text{Runset}| \ge m$. If $|\mathcal{S}| < m$, all available workloads will be executed regardless of imbalance. We suggest the value of $m$ to be $0.5\sim 0.75\times$ of the number of micro-batches, according to empirical evidences in \cref{subsec:ablation_micro_batching}.

EasyBalance provides instant flexibility across tasks, as scheduling decisions are made solely based on current workload distributions and do not rely on task-specific routing statistics. Moreover, it incurs negligible additional overhead, as no expert replication or migration is involved, addressing the aforementioned limitation on scalability. The scheduling algorithm operates only on small-sized routing metadata, and its runtime cost is lightweight and negligible compared to the end-to-end inference latency (see empirical results in \cref{tab:scheduling_latency}).

\section{Experiments}

\textbf{Models}. We evaluate EasyBalance on 3 representative open-source MoE models: Qwen3-30B-A3B-Instruct-2507 (Qwen3-30B) \cite{qwen3}, Moonlight-16B-A3B-Instruct (Moonlight-16B) \cite{moonlight} and Qwen3-235B-A22B-Instruct-2507 (Qwen3-235B). Except for results in \cref{subsec:combination_to_EPLB}, we use the standard expert-device mapping where experts are evenly distributed across devices. The number of activated experts per token follows the default value in model configurations. 

\textbf{Benchmark}. We use LongBench \cite{longbench} as the evaluation benchmark. LongBench is designed for large-scale and long-context inference, covering a diverse set of tasks including passage comprehension, question answering, information retrieval, summarization, and code generation. This task diversity induces highly heterogeneous routing patterns across inputs, making LongBench well suited for evaluating the robust flexibility of methods.

\textbf{Configurations}. All experiments are conducted on a single node equipped with 8$\times$A800-SXM4 GPUs 80GB, interconnected via NVLink. Unless otherwise stated, the expert parallelism size is set to 8. The batch sizes and token sequence lengths of each GPU are set to 16*4K and 12*512 for smaller (Qwen3-30B and Moonlight-16B) and larger (Qwen3-235B) models respectively. The number of micro-batch splits is 4 with $m$ set to 3, as this setting typically achieves optimal performances.

\textbf{Metrics}. We evaluate inference speed with end-to-end latency, which directly reflects the empirical performances. Additionally, we report effective workload (defined in \cref{eq:effective_workload_def}) and GPU utilization (defined in \cref{eq:utilization_def}) as quantitative indicators of load imbalance. For readability, we report GPU ``under-utilization'' as $1-u$. The latencies are averaged over 8 runs to minimize the impact of system fluctuations.

\begin{figure*}[tp]
    \centering
    \begin{subfigure}[tb]{0.7\textwidth}
        \centering
        \includegraphics[width=\textwidth]{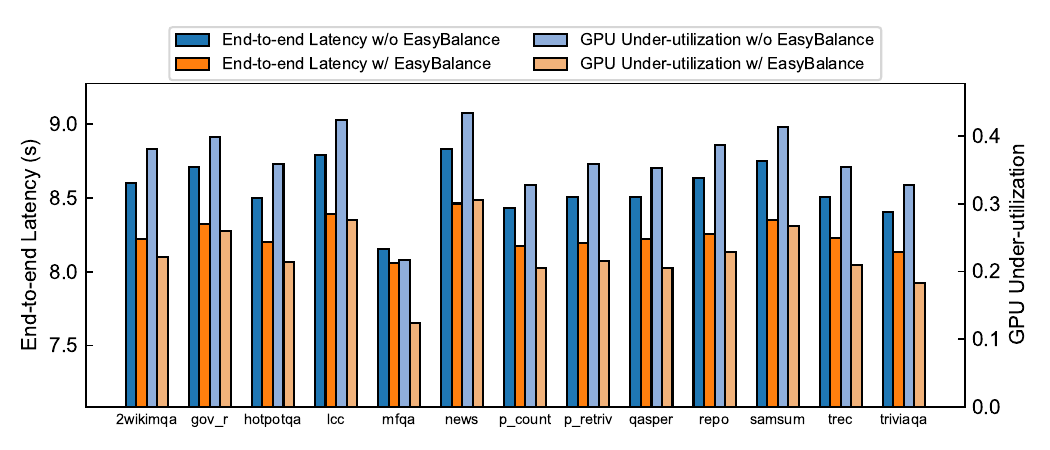}
        \caption{Qwen3-30B}
        \label{subfig:main_result_qwen3_30b}
    \end{subfigure}
    \hfill
    % Subfigure 2
    \begin{subfigure}[tb]{0.7\textwidth}
        \centering
        \includegraphics[width=\textwidth]{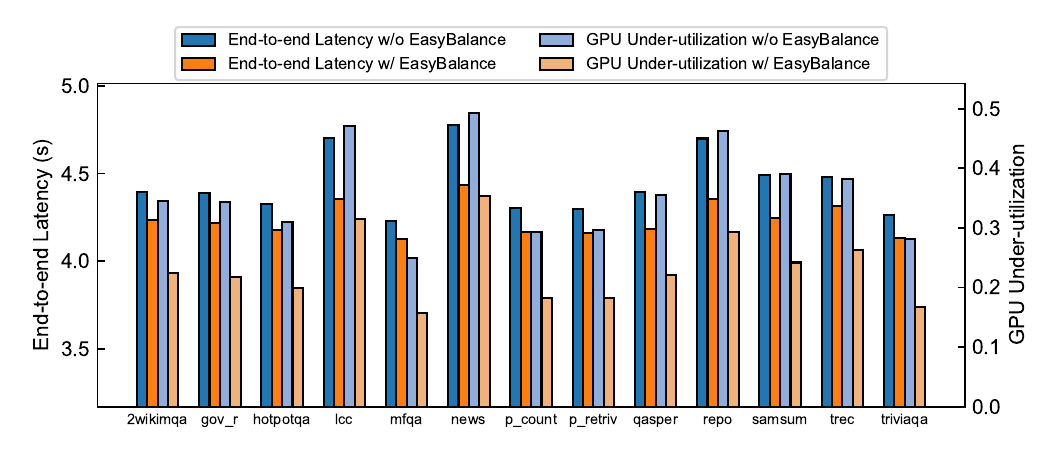}
        \caption{Moonlight-16B}
        \label{subfig:main_result_moonlight_16b}
    \end{subfigure}
    % Subfigure 3
    \begin{subfigure}[tb]{0.7\textwidth}
        \centering
        \includegraphics[width=\textwidth]{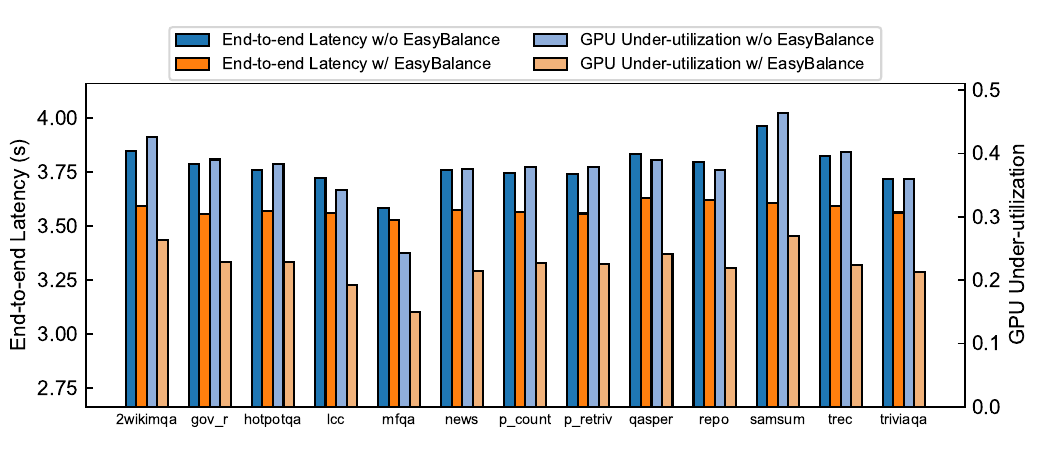}
        \caption{Qwen3-235B}
        \label{subfig:main_result_qwen3_235b}
    \end{subfigure}

\caption{End-to-end latency(s) and GPU under-utilization across tasks and models. Lower is better.}
\label{fig:main_results}
\end{figure*}

\begin{figure}[t]
    \centering
    \begin{subfigure}{\linewidth}
        \includegraphics[width=\linewidth]{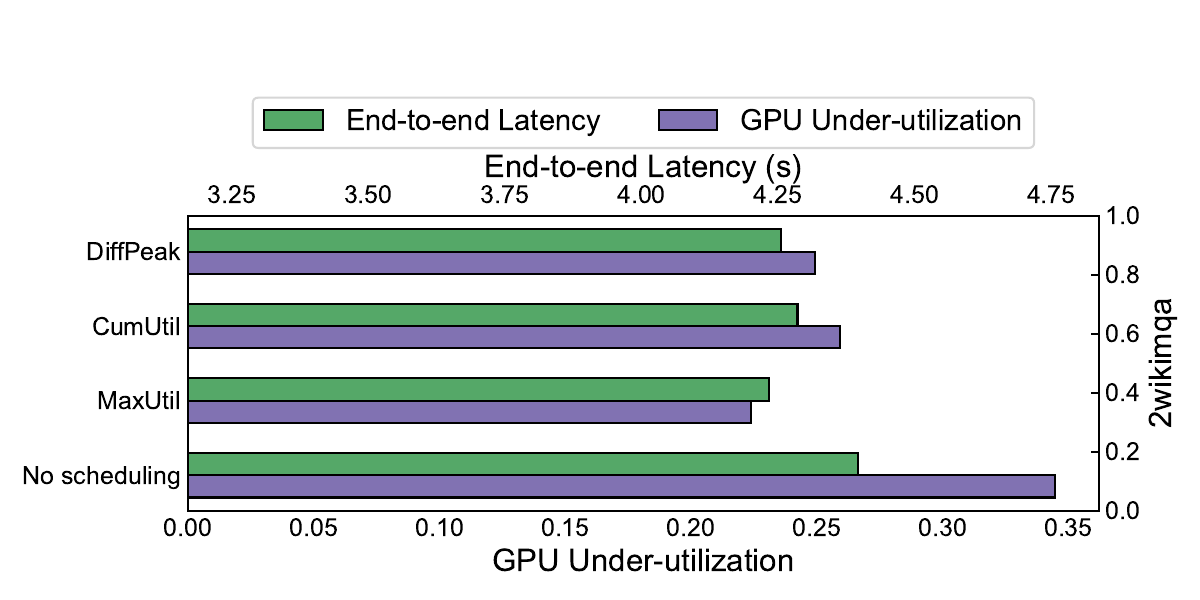}
        \caption{2wikimqa}
    \end{subfigure}
    \begin{subfigure}{\linewidth}
        \includegraphics[width=\linewidth]{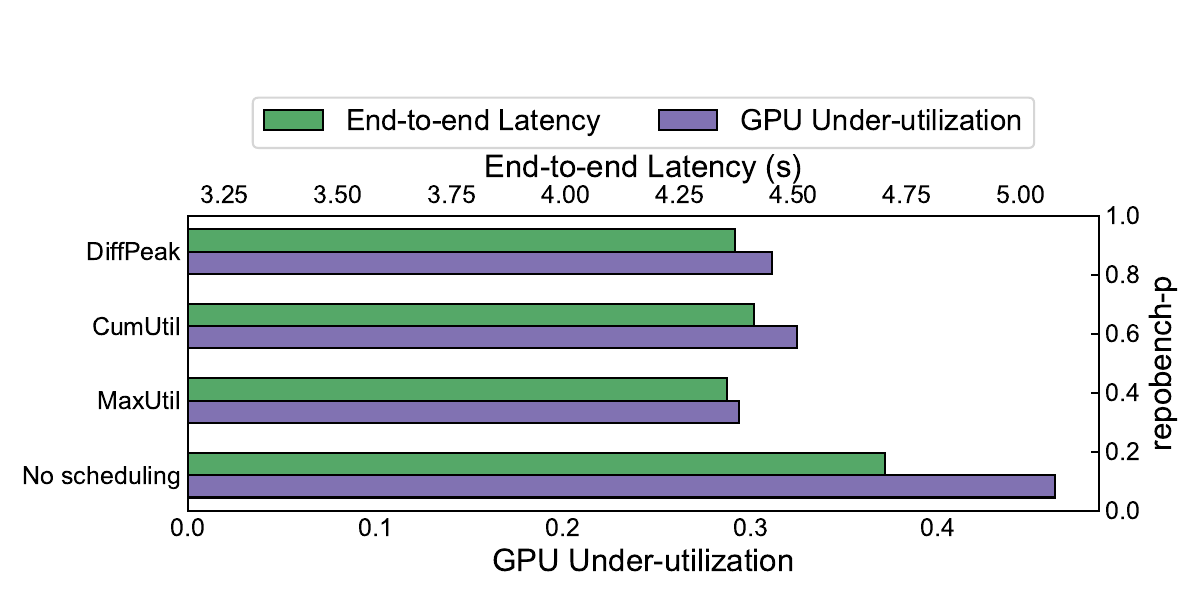}
        \caption{repobench-p}
    \end{subfigure}
    \caption{End-to-end latency(s) and GPU under-utilization of different scheduling algorithms, applied to Moonlight-16B on representative tasks. Lower is better.}
    \label{fig:strategies}
\end{figure}

\subsection{Main Results}

\cref{fig:main_results} presents the end-to-end latencies and GPU under-utilization of tested models across various tasks. As shown, EasyBalance consistently improves inference-speed performance across models and task categories, significantly alleviating GPU under-utilization caused by expert load imbalance by mostly more than 40\% ($\ge$0.35 to $\approx$0.2). The consistent effectiveness across tasks demonstrates strong robustness and flexibility of our method, which is attributed to the design that does not require expert-mapping modifications.

\begin{figure}[t]
    \centering
    \includegraphics[width=\linewidth]{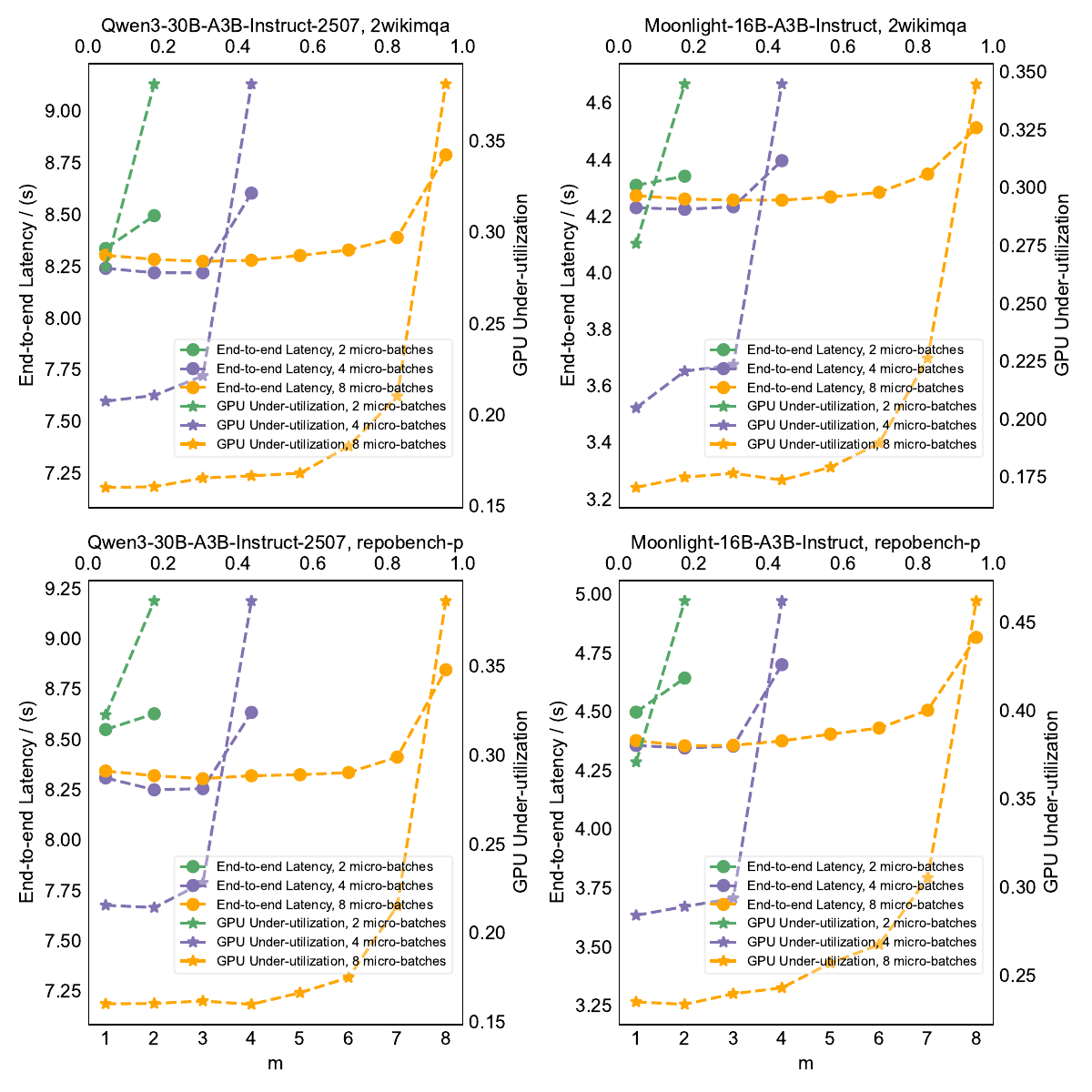}
    \caption{End-to-end latency(s) and GPU under-utilization of different micro-batches and $m$ on representative tasks. Results are the best among scheduling algorithms in \cref{subsec:scheduling_alg}. Lower is better.}
    \label{fig:micro_batches}
\end{figure}

\begin{table}[t]
    \centering
    \begin{tabular}{c|ccc}
    \toprule
       ~ & MaxUtil & CumUtil & DiffPeak  \\
    \midrule
       Scheduling(ms) & 0.277 & 0.267 & 0.130 \\ % 4
       \midrule
       End-to-end(s) & 8.21 & 8.27 & 8.24 \\
    \bottomrule
    \end{tabular}
    \caption{Scheduling latency(ms) per step and end-to-end latency(s) with different strategies upon Qwen3-30B and 2wikimqa. The batch size and sequence length are 128 and 4K, and the number of micro-batches is 4.}
    \label{tab:scheduling_latency}
\end{table}

\subsection{Scheduling Algorithms}
\label{subsec:scheduling_alg}

The scheduling procedure incurs only negligible overhead, as shown in \cref{tab:scheduling_latency}. However, while the utilization-maximization (\textbf{MaxUtil}) scheduling strategy generally achieves best performances, its overhead grows exponentially with the number of batches ($\mathrm{}{O}(2^n)$ with batch size $n$). Therefore, we additionally evaluate two alternative heuristic algorithms whose overhead scales linearly ($\mathrm{}{O}(n)$).

Denote the workload distribution of micro-batch $i$ across $D$ devices as $\boldsymbol{w}_i=(w^0_i, w^1_i, \dots, w^{D-1}_i)$, the algorithms are:

\textbf{Cumulated Utilization (CumUtil)}. A batch will be scheduled only if its inclusion improves the overall GPU utilization. The algorithm evaluates batches sequentially from $0$ to $N-1$, with the first batch always scheduled. Formally, at step $i$, given the currently chosen workload set $S_{i,j}$, batch $j+1$ is added to form $S_{i,j+1}$ if
\[
U(S_{i,j}\cup\{\boldsymbol{w}_{i,j+1}\}) > U(S_{i,j}),
\]
where $U(\cdot)$ denotes GPU utilization. Otherwise $S_{i,j+1}=S_{i,j}$. $S_{i,0}=\{0\}$.

\textbf{Different Peak Devices (DiffPeak)}. A batch will be scheduled only when the peak device of its workload differs from those of all chosen micro-batches. Formally, at step $i$, batch $j+1$ is scheduled if
\[
\text{argmax}_d(\boldsymbol{w}_{i,j+1}) \notin \{\text{argmax}_d(\boldsymbol{w}_{i,k})|\boldsymbol{w}_{i,k} \in S_{i,j}\}.
\]

Compared to the original algorithm, these two alternatives do not account for the interaction between workloads across micro-batches, but only each independent workload. The results in \cref{fig:strategies} show that, while these alternatives underperform the MaxUtil strategy, they both outperform the vanilla baseline (i.e., no cross-layer combination), suggesting that the robust effectiveness of EasyBalance does not critically depend on a specific scheduling heuristic. 

\begin{table}[t]
    \centering
    \begin{tabular}{c|c|ccc}
    \toprule
\multirow{2}{*}{Task} & \multirow{2}{*}{Method} & \multicolumn{3}{c}{Expert-Parallel Size} \\
        \cmidrule(lr){3-5}
    ~ & ~ & 2  & 4 & 8  \\
    \midrule
    \multicolumn{5}{c}{Qwen3-30B} \\
    \midrule
    \multirow{2}{*}{2wikimqa} & w/o EB & 0.10 & 0.21  & 0.37 \\
    ~ & w/ EB & 0.04 & 0.12 & 0.21 \\
    \midrule
    \multirow{2}{*}{repobench-p} & w/o EB & 0.10 & 0.24  & 0.38 \\
    ~ & w/ EB & 0.04 & 0.12 & 0.23 \\
    \midrule
    \multirow{2}{*}{trec} & w/o EB & 0.08 & 0.20  & 0.35 \\
    ~ & w/ EB & 0.04 & 0.10 & 0.21 \\
    \midrule
    \multicolumn{5}{c}{Moonlight-16B} \\
    \midrule
    \multirow{2}{*}{2wikimqa} & w/o EB & 0.06 & 0.18  & 0.33 \\
    ~ & w/ EB & 0.03 & 0.11 & 0.21 \\
    \midrule
    \multirow{2}{*}{repobench-p} & w/o EB & 0.12 &  0.29 & 0.46 \\
    ~ & w/ EB & 0.05  & 0.17 & 0.29 \\
    \midrule
    \multirow{2}{*}{trec} & w/o EB & 0.07 & 0.21  & 0.38 \\
    ~ & w/ EB & 0.04 & 0.13 & 0.26 \\    
    \bottomrule
    \end{tabular}
    \caption{GPU under-utilization of Qwen3-30B and Moonlight-16B on representative tasks. The batch size for EP=2,4,8 is 32,64,128 respectively. EB stands for EasyBalance.}
    \label{tab:ablation_ep_size}
    % \vspace{-10pt}
\end{table}

\subsection{Orthogonality to EPLB}
\label{subsec:combination_to_EPLB}

\begin{figure*}[t]
    \centering
    \includegraphics[width=0.7\linewidth]{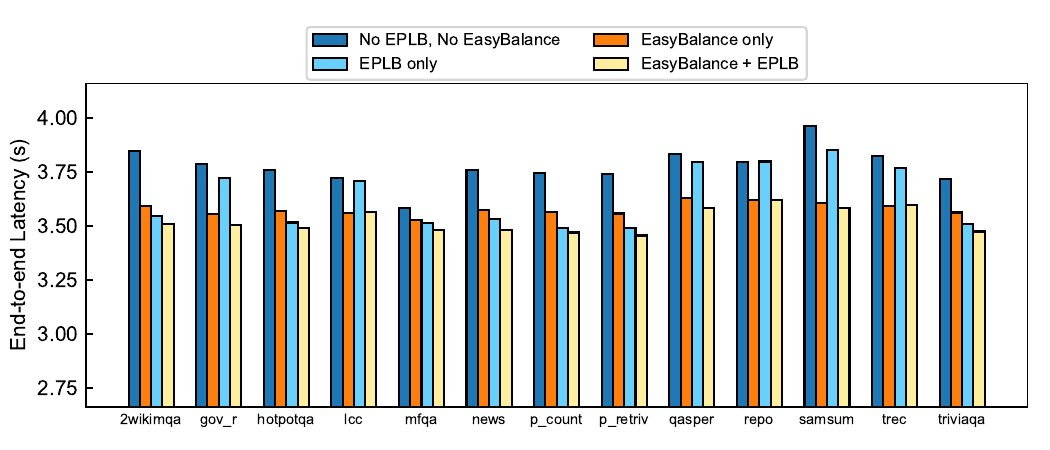}
    \caption{End-to-end latency(s) and GPU under-utilization with/without EPLB on Qwen3-235B. The EPLB expert placement is generated from the routing distribution of all 13 representative tasks of LongBench, with 128 physical experts.}
    \label{fig:orthogonality_with_EPLB}
\end{figure*}

\begin{figure*}[t]
    \centering
    \includegraphics[width=0.75\textwidth]{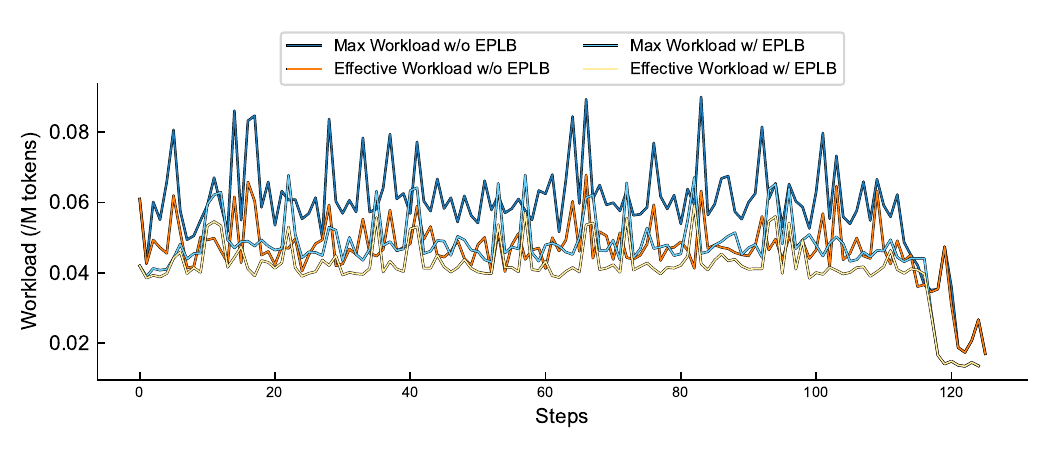}
    \caption{Maximum and effective workloads(/M tokens) of Qwen3-235B on multi\_news in one run. The EPLB expert placement is generated from the routing distributions of all 13 representative tasks of LongBench, with 128 physical experts.}
    \label{fig:eplb_step_plot}
\end{figure*}

EasyBalance is orthogonal to load balancing methods that require expert-map modification, as it is agnostic to the mapping itself. To validate this, we evaluate EasyBalance in combination with EPLB \cite{eplb}. We configure EPLB’s expert placements using the routing history from all 13 representative tasks. The results in \cref{fig:orthogonality_with_EPLB} show that EasyBalance consistently provides additive performance gains, confirming the orthogonality.

\cref{fig:eplb_step_plot} shows the maximum and effective workloads of each scheduling step, where the two metrics are defined as $\sum_j\max(\boldsymbol{w}_{i,j})$ and $\max\sum_j \boldsymbol{w}_{i,j}$ respectively. The results show that EasyBalance effectively reduces effective workloads in most cases through cross-layer balancing opportunities ($\sum_j\max(\boldsymbol{w}_{j,i})<\max\sum_j \boldsymbol{w}_{j,i}$).

Notably, the performance gains of EPLB differ substantially across task types, highlighting the task-dependent nature of its gains and therefore the limited flexibility of existing map-modification methods. In contrast, EasyBalance achieves consistent and stable effectiveness across all evaluated tasks.

\subsection{Ablations}

The ablation studies are about configurations of micro-batching and expert parallelism.

\subsubsection{Micro-Batching}
\label{subsec:ablation_micro_batching}

We vary both the number of micro-batches and the minimum execution threshold $m$. The results in \cref{fig:micro_batches} demonstrate that the effectiveness of EasyBalance remains consistent across different configurations. The value of $m$ plays a critical role in the scheduling behavior: a smaller $m$ may restrict the scheduler to explore more balancing opportunities, while a larger one may force imbalanced workloads to be scheduled. Based on the empirical results, we suggest the value of $m$ to be $0.5\sim 0.75\times$ of the number of micro-batches.

Choosing an appropriate micro-batch size is essential for achieving optimal inference performance. Excessively small sizes (e.g., 2) result in insufficient imbalance mitigation, whereas overly large sizes reduce computation intensity and cause slowdown \cite{speculative_decoding}. Notably, the optimal performance is achieved with 4 micro-batches, while the highest GPU utilization occurs at a size of 8. This suggests the potential of even greater performance gains that can be realized under heavier or more imbalanced inference workloads.

\subsubsection{Parallelism Configurations}

\cref{tab:ablation_ep_size} reports the GPU under-utilization of various expert-parallel (EP) configurations. To ensure a fair comparison, we proportionally adjust the global batch size so that the per-device capacity remains constant (e.g., 32, 64, 128 for EP size = 2, 4, 8 respectively). The results indicate that EasyBalance consistently reduces the effective workloads across all EP configurations.

As demonstrated, larger EP sizes generally result in lower GPU utilization, which is consistent with prior studies \cite{large_scale_expert_parallelism, eplb}. This is because co-located experts can absorb each other's workload imbalance within a device. As the EP size increases, each device hosts fewer experts, reducing local absorbing opportunities before routing skew translates into inter-device imbalance. Therefore, the limitation on scalability is crucial for load-balancing methods, while EasyBalance is designed to address this challenge.

\section{Related Works}

\textbf{Distributed MoE Inference}. Existing works focus on various aspects of distributed MoE inference acceleration. ScMoE \cite{shortcut} adopts a shortcut connection to break layerwise dependencies between computation and communication. Tutel \cite{tutel} designs an identical distribution layout of MoE models for switchable parallelism and dynamic pipelining. KTransformers \cite{ktransformers} develops faster CPU kernels and proposes expert deferral for device-heterogeneous distributed MoE inference. DeepSpeed-MoE \cite{deepspeed-moe} implements hierarchical all-to-all based on tensor- and expert-parallel topologies to reduce communication overheads. Occult \cite{occult} combines co-activated experts onto the same device for efficient all-to-all communication. 

\textbf{Expert-Parallel Load Balancing}. Existing works on EP load balancing mainly address the problem through routing statistic analysis, expert replication, and expert migration. Lina \cite{lina} observes that tokens routed to the same expert in one layer are more likely to be routed to another same expert in the following layer, and exploits this property to profile routing patterns and guide expert replication. EPLB \cite{eplb} introduces expert redundancy by creating replicated experts and dynamically adjusting expert placement based on historical routing information. Harmony \cite{harmoeny} further adapts expert workload allocation according to current routing results, while migrating experts in the background of computation to reduce the impact of additional overhead.

Existing load-balancing strategies are largely constrained in flexibility and scalability, due to their reliance on modifying expert-device mappings. EasyBalance addresses this limitation by exploring an orthogonal direction: expert-map-independent load balancing, which has been largely unexplored in prior studies.

\section{Conclusion}

This paper proposes EasyBalance, a cross-layer load-balancing strategy for expert-parallel MoE inference. The key insights are cross-layer expert redundancy and workload combination. Without requiring expert replication or migration, EasyBalance creates load-balancing opportunities by scheduling micro-batches to execute MoE computation across different layers. EasyBalance consistently accelerates distributed MoE inference, improving GPU utilization by mostly over 40\%.

\clearpage
\section*{Acknowledgments}

This work is partially supported by the NSF of China (under Grant 92364202), and Major Program of ISCAS (Grant No. ISCAS-ZD-202402).

\section*{Impact Statement}

This paper presents work whose goal is to advance the field of machine learning. There are many potential societal consequences of our work, none of which we feel must be specifically highlighted here.


\bibliography{paper}
\bibliographystyle{icml2026}

%%%%%%%%%%%%%%%%%%%%%%%%%%%%%%%%%%%%%%%%%%%%%%%%%%%%%%%%%%%%%%%%%%%%%%%%%%%%%%%
%%%%%%%%%%%%%%%%%%%%%%%%%%%%%%%%%%%%%%%%%%%%%%%%%%%%%%%%%%%%%%%%%%%%%%%%%%%%%%%
% APPENDIX
%%%%%%%%%%%%%%%%%%%%%%%%%%%%%%%%%%%%%%%%%%%%%%%%%%%%%%%%%%%%%%%%%%%%%%%%%%%%%%%
%%%%%%%%%%%%%%%%%%%%%%%%%%%%%%%%%%%%%%%%%%%%%%%%%%%%%%%%%%%%%%%%%%%%%%%%%%%%%%%
\newpage
\appendix
\onecolumn

\section{Workload Distributions across Tasks}

\begin{figure*}[h]
    \centering
    \includegraphics[width=0.65\linewidth]{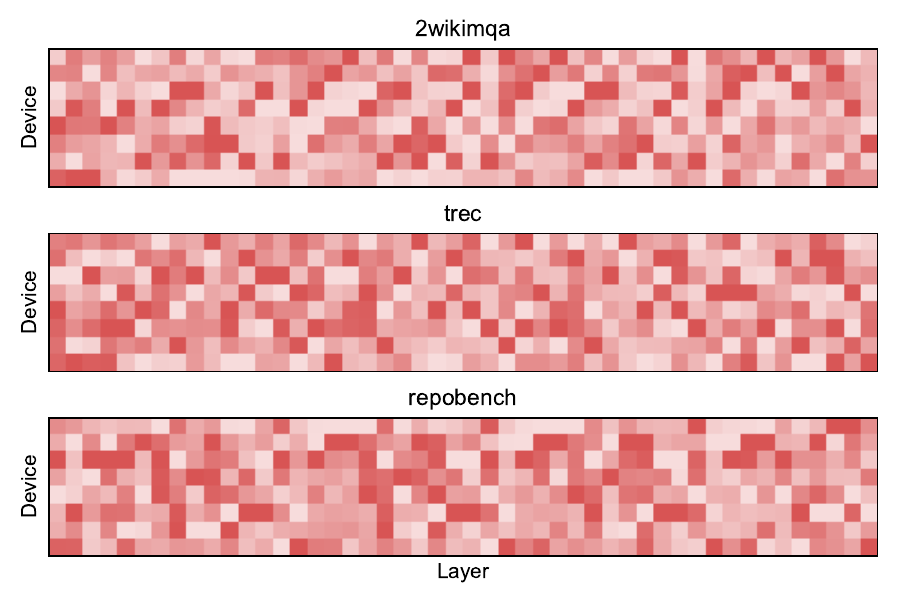}
    \caption{Workload distributions of Qwen3-30B on representative tasks under EP=8. Lighter colors indicate lower workloads.}
    \label{fig:routing_distribution}
\end{figure*}

\cref{fig:routing_distribution} demonstrates the routing pattern of each layer of Qwen3-30B models, with lighter colors indicating lower workloads. According to the results, the routing distributions are highly task-dependent, posing limitations upon flexibility of load balancing methods that require expert-map modification.

\section{Clarification of Figures}

We report end-to-end latencies as the performance metric, yet EasyBalance targets only at MoE computation, rather than attention. Formally, we have

$$t_{e2e} = t_{expert} + t_{attention}$$

Note that under the same settings of batch size and sequence length, $t_{attention}$ is constant across different tasks (irrelevant to MoE imbalance). Therefore, we set the start of y-axis of figures to be $t_{attention}$ for better demonstration of ``expert latency''.

\section{More Results}

\subsection{EPLB with Redundant Experts}
\label{subsec:eplb_with_redundant_experts}

EPLB supports redundant expert placement, where redundant experts are replications of some hot experts from other EP ranks. However, the results in \cref{tab:workload_various_global_experts} show that introducing redundant experts yields only marginal improvements on load balancing, while incurring substantial memory overhead. We evaluate Qwen3-30B with 128 logical experts under EP=8 and observe that the utilization improvement for 136 physical experts (8 replicas, 1 for each EP rank) is marginal, especially compared to the significant effectiveness of EasyBalance. The results also show that EasyBalance is orthogonal to expert replication, while incurring no memory overhead.

\begin{table}[htb]
    \centering
    \begin{tabular}{c|cccc}
    \toprule
      Task   &  EPLB(128) & EPLB(128)+EasyBalance & 
      EPLB(136) & EPLB(136)+EasyBalance \\
    \midrule
    2wikimqa  &  0.23 & \textbf{0.14} & 0.22 & \textbf{0.12} \\
    trec  &  0.29 & \textbf{0.16} & 0.28 & \textbf{0.17} \\
    \bottomrule
    \end{tabular}
    \caption{GPU under-utilization of EPLB with various numbers of physical experts on Qwen3-30B. The batch size and sequence length are 128 and 4K, respectively.}
    \label{tab:workload_various_global_experts}
\end{table}

\subsection{Scheduling Algorithms}

\begin{figure}[t]
    \centering
    \newcommand{\subfigwidth}{0.48\textwidth}
    \begin{subfigure}{\subfigwidth}
        \centering
        \includegraphics[width=\linewidth]{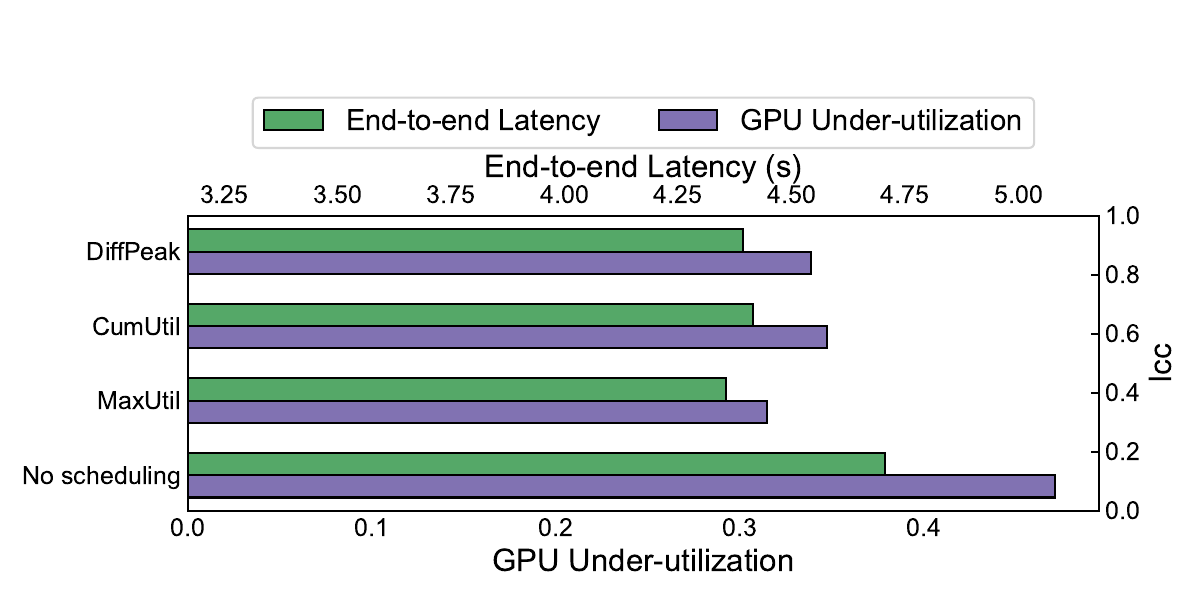}
        \caption{Moonlight-16B, lcc}
    \end{subfigure}
    \hfill
    \begin{subfigure}{\subfigwidth}
        \centering
        \includegraphics[width=\linewidth]{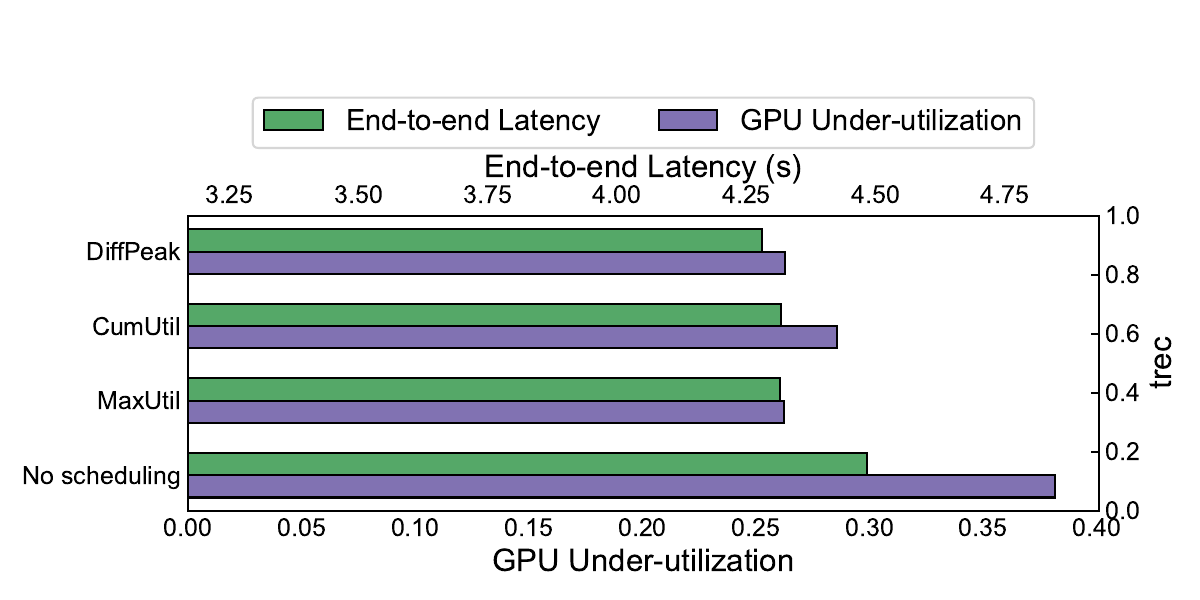}
        \caption{Moonlight-16B, trec}
    \end{subfigure}
    
    \begin{subfigure}{\subfigwidth}
        \centering
        \includegraphics[width=\linewidth]{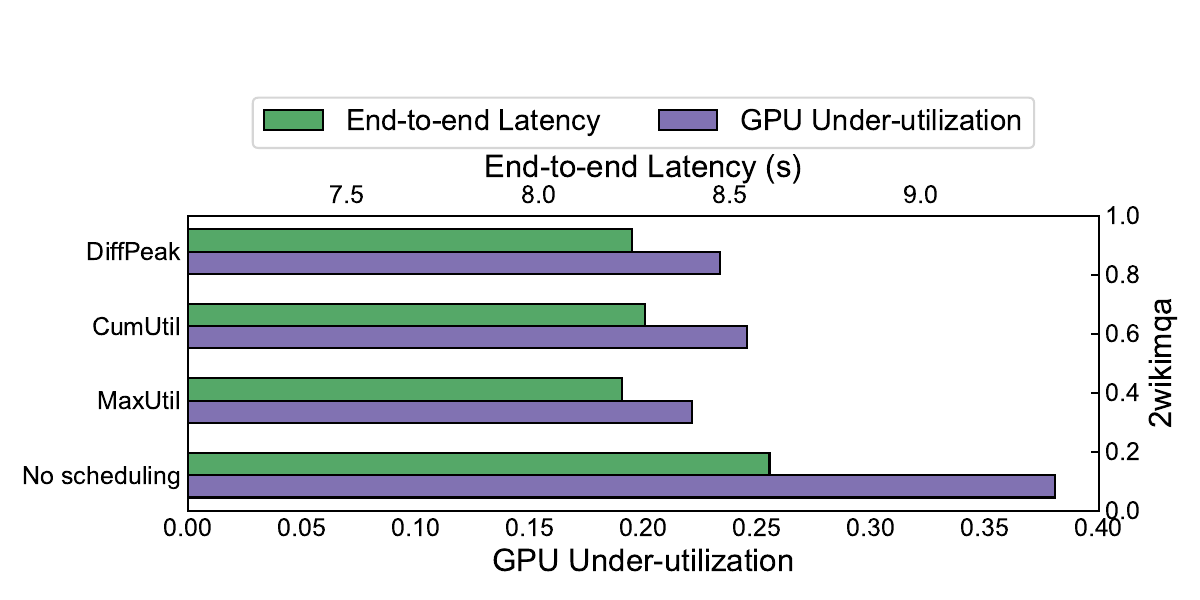}
        \caption{Qwen3-30B, 2wikimqa}
    \end{subfigure}
    \hfill
    \begin{subfigure}{\subfigwidth}
        \centering
        \includegraphics[width=\linewidth]{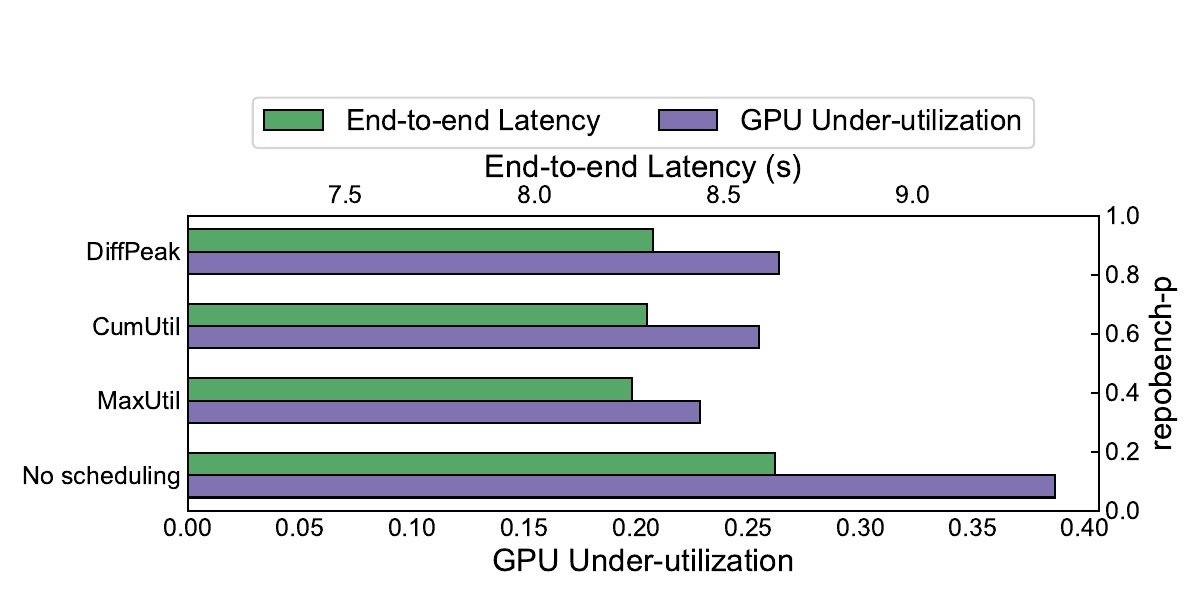}
        \caption{Qwen3-30B, repobench-p}
    \end{subfigure}

    \begin{subfigure}{\subfigwidth}
        \centering
        \includegraphics[width=\linewidth]{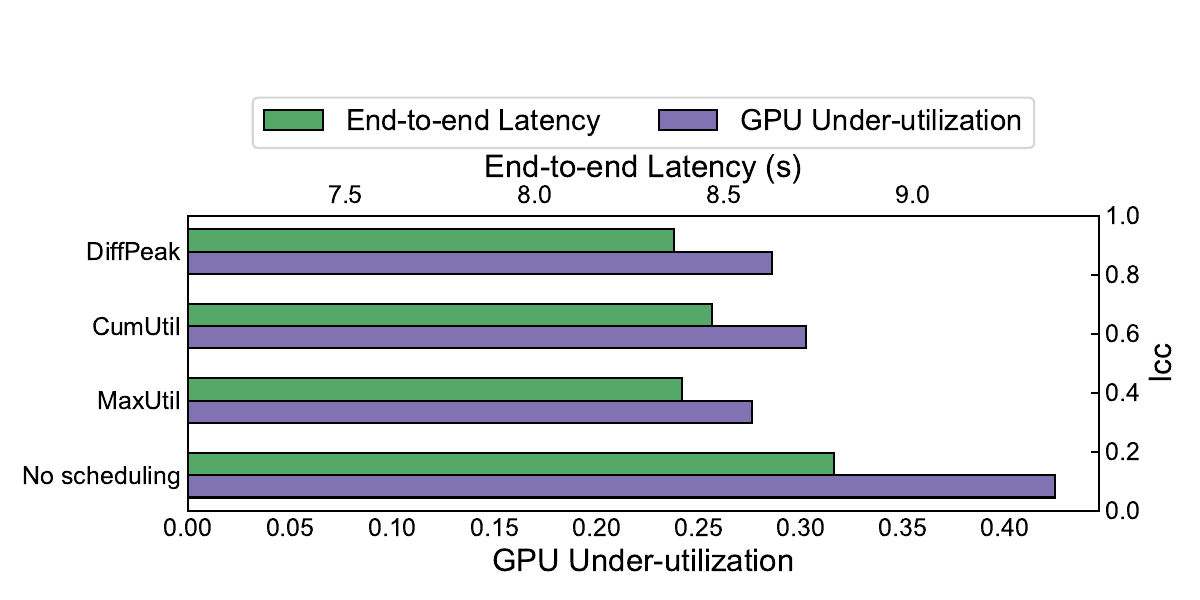}
        \caption{Qwen3-30B, lcc}
    \end{subfigure}
    \hfill
    \begin{subfigure}{\subfigwidth}
        \centering
        \includegraphics[width=\linewidth]{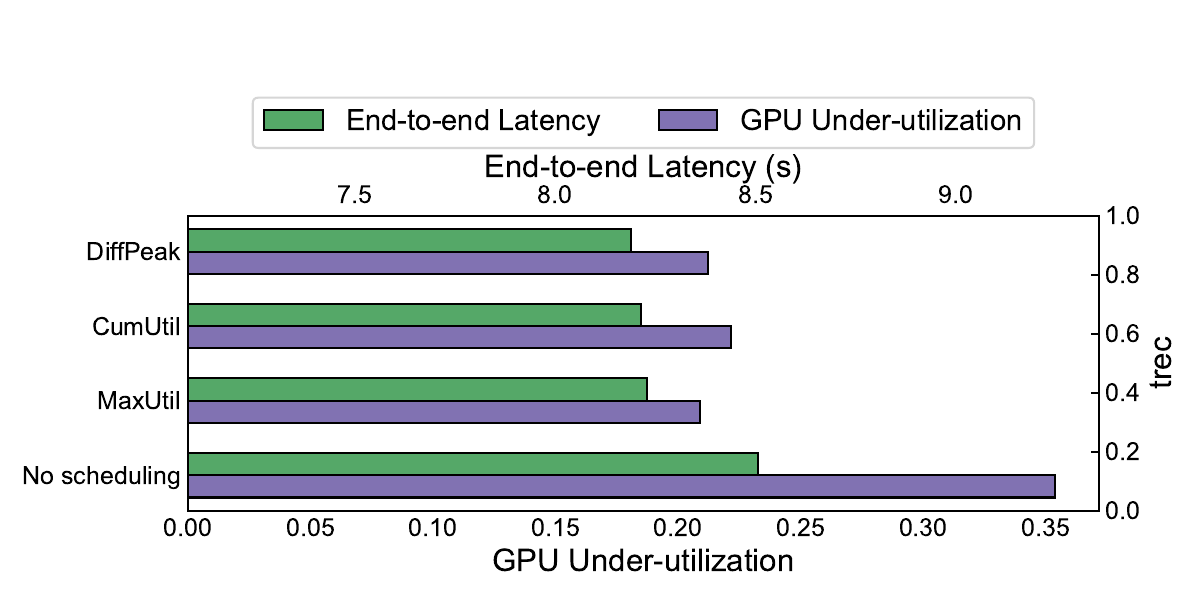}
        \caption{Qwen3-30B, trec}
    \end{subfigure}

    \caption{End-to-end latency(s) and GPU under-utilization of different scheduling algorithms across tasks and models.}
    \label{fig:more_scheduling_algorithms}
\end{figure}

\cref{fig:more_scheduling_algorithms} shows more results about different scheduling algorithms. The results indicate that MaxUtil may underperform other strategies in some cases, yet all the algorithms effectively reduce end-to-end latency and GPU under-utilization.

\subsection{Effectiveness of Cross-Layer Workload Combination}

\begin{figure}[p]
    \centering
    \newcommand{\subfigwidth}{0.7\textwidth}
    \begin{subfigure}{\subfigwidth}
        \centering
        \includegraphics[width=\linewidth]{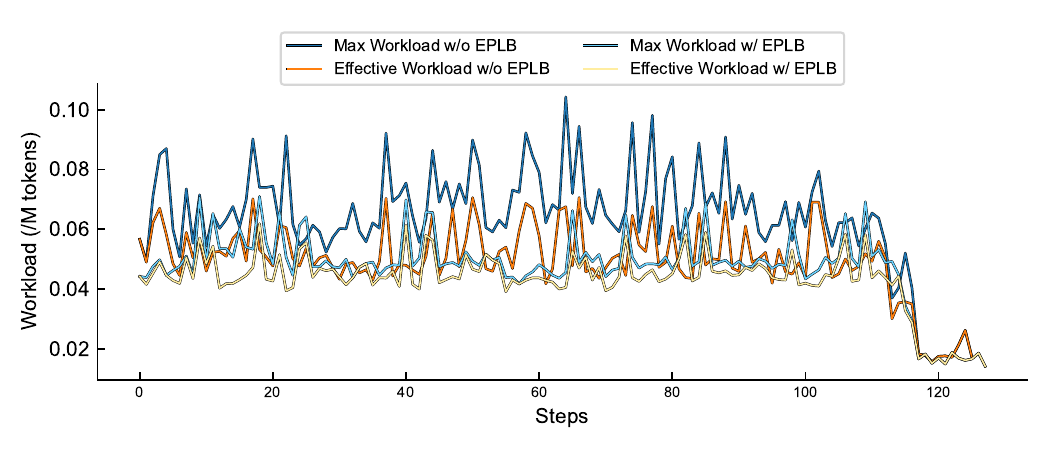}
        \caption{2wikimqa}
    \end{subfigure}
    \begin{subfigure}{\subfigwidth}
        \centering
        \includegraphics[width=\linewidth]{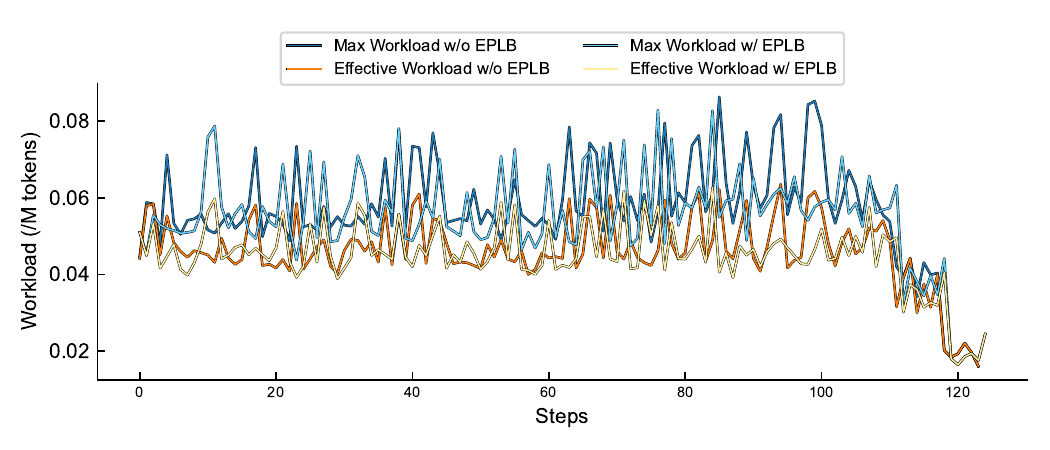}
        \caption{lcc}
    \end{subfigure}

    \begin{subfigure}{\subfigwidth}
        \centering
        \includegraphics[width=\linewidth]{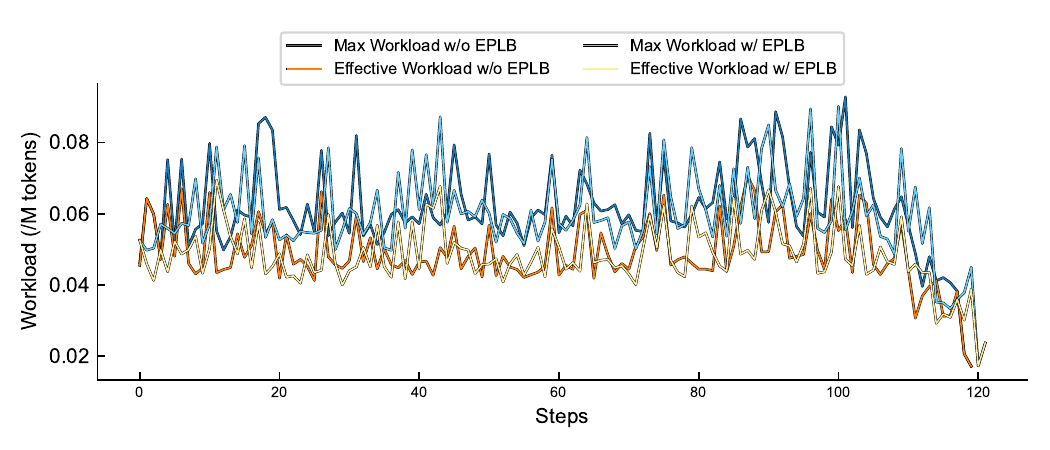}
        \caption{repobench-p}
    \end{subfigure}

    \begin{subfigure}{\subfigwidth}
        \centering
        \includegraphics[width=\linewidth]{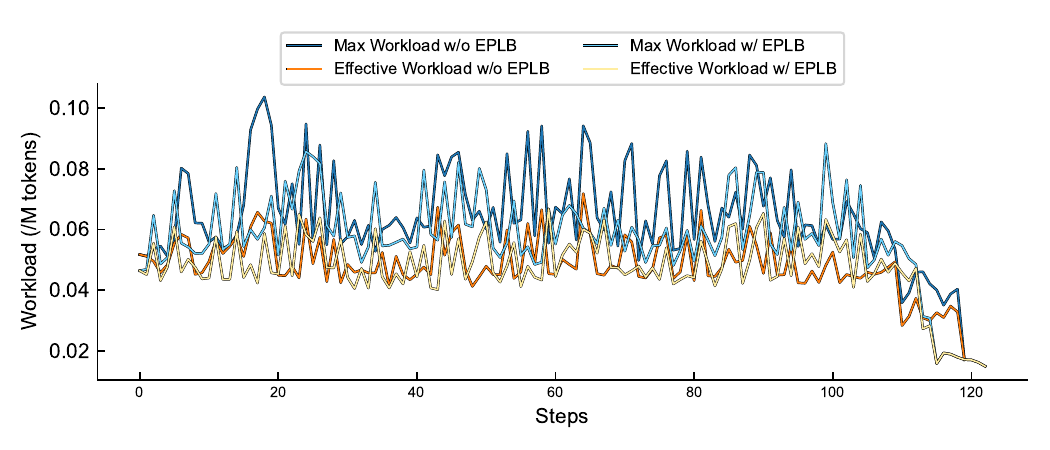}
        \caption{trec}
    \end{subfigure}

    \caption{Maximum and effective workloads of Qwen3-235B across tasks in one run. The EPLB expert placement is the same as \cref{fig:eplb_step_plot}.}
    \label{fig:more_eplb_step_plot}
\end{figure}

\cref{fig:more_eplb_step_plot} shows additional results of per-step workload statistics upon Qwen3-235B across more tasks, as a complementary to \cref{fig:eplb_step_plot}. Results show that workloads can mitigate each other's imbalance in most cases, demonstrating the effectiveness of our method.

\section{Limitations}

EasyBalance is effective under distributed MoE inference , while it has no acceleration on non-MoE models or no-EP scenarios. EasyBalance currently requires micro-batching, while the potential of applying it to a single sequence (by splitting the sequence into pieces and run them on different layers) remain unexplored. The experiments are conducted only on A800-SXM4 80GB GPUs, leaving its empirical effectiveness on other platforms uncertain. EasyBalanec is potentially effective under Attention-FFN Disaggregation \cite{step3}, while the effectiveness remains unverified.

%%%%%%%%%%%%%%%%%%%%%%%%%%%%%%%%%%%%%%%%%%%%%%%%%%%%%%%%%%%%%%%%%%%%%%%%%%%%%%%
%%%%%%%%%%%%%%%%%%%%%%%%%%%%%%%%%%%%%%%%%%%%%%%%%%%%%%%%%%%%%%%%%%%%%%%%%%%%%%%

\end{document}